%% file: main.tex
\documentclass[runningheads]{llncs}

\usepackage{eccv}

\usepackage{eccvabbrv}

\usepackage{graphicx}
\usepackage{booktabs}

\usepackage[accsupp]{axessibility}

\input{preamble}
\usepackage[pagebackref,breaklinks,colorlinks,citecolor=eccvblue]{hyperref}
\usepackage{hyperref}

\usepackage{orcidlink}

\begin{document}

\title{Uncertainty Calibration with Energy Based Instance-wise Scaling in the Wild Dataset}

\titlerunning{Uncertainty Calibration with Energy Based Instance-wise Scaling}

\author{Mijoo Kim\inst{1}\orcidlink{0000-0002-0397-1852} \and
Junseok Kwon\inst{1}\orcidlink{0000-0001-9526-7549}}

\authorrunning{M.~Kim et al.}

\institute{Chung-Ang University, Seoul, Korea
\\
\email{\{mijoo707,jskwon\}@cau.ac.kr}}

\maketitle

\input{sec/0_abstract}

\input{sec/1_introduction}
\input{sec/2_related_work}

\input{sec/3_problem_setup}
\input{sec/4_proposed_method}
\input{sec/5_experiment}
\input{sec/6_conclusion}

\bibliographystyle{splncs04}
\bibliography{main}

\newpage
\input{sec/7_supp}

\end{document}

%% file: preamble.tex
\usepackage[dvipsnames]{xcolor}

\usepackage{algorithm}
\usepackage{algorithmic}
\usepackage{amsfonts}
\usepackage{amsmath}
\usepackage{soul}
\usepackage{multirow}
\usepackage{xcolor}
\usepackage{color}
\usepackage{colortbl,hhline}
\definecolor{Gray}{gray}{0.95}

\usepackage{adjustbox}

\usepackage{rotfloat,array}
\newcolumntype{C}[1]{>{\centering\arraybackslash}p{#1}}
\newcolumntype{L}[1]{>{\arraybackslash}m{#1}}

\newcommand{\dtoprule}{\specialrule{1pt}{0pt}{0.4pt}%
            \specialrule{0.3pt}{0pt}{\belowrulesep}%
            }
\newcommand{\dbottomrule}{\specialrule{0.3pt}{0pt}{0.4pt}%
            \specialrule{1pt}{0pt}{\belowrulesep}%
            }

%% file: sec/0_abstract.tex
\begin{abstract}
With the rapid advancement in the performance of deep neural networks (DNNs), there has been significant interest in deploying and incorporating artificial intelligence (AI) systems into real-world scenarios. However, many DNNs lack the ability to represent uncertainty, often exhibiting excessive confidence even when making incorrect predictions.
To ensure the reliability of AI systems, particularly in safety-critical cases, DNNs should transparently reflect the uncertainty in their predictions. In this paper, we investigate robust post-hoc uncertainty calibration methods for DNNs within the context of multi-class classification tasks. While previous studies have made notable progress, they still face challenges in achieving robust calibration, particularly in scenarios involving out-of-distribution (OOD).
We identify that previous methods lack adaptability to individual input data and struggle to accurately estimate uncertainty when processing inputs drawn from the wild dataset. To address this issue, we introduce a novel instance-wise calibration method based on an energy model. Our method incorporates energy scores instead of softmax confidence scores, allowing for adaptive consideration of DNN uncertainty for each prediction within a logit space.
In experiments, we show that the proposed method consistently maintains robust performance across the spectrum, spanning from in-distribution to OOD scenarios, when compared to other state-of-the-art methods. The source code is available at  \href{https://github.com/mijoo308/Energy-Calibration}{https://github.com/mijoo308/Energy-Calibration}.

\keywords{Uncertainty Calibration \and Out-of-distribution \and Energy based
instance-wise scaling}

\end{abstract}

%% file: sec/1_introduction.tex
\section{Introduction}
\label{sec:intro}

\begin{figure}[t]
\centering
    \includegraphics[width=1.0\columnwidth]{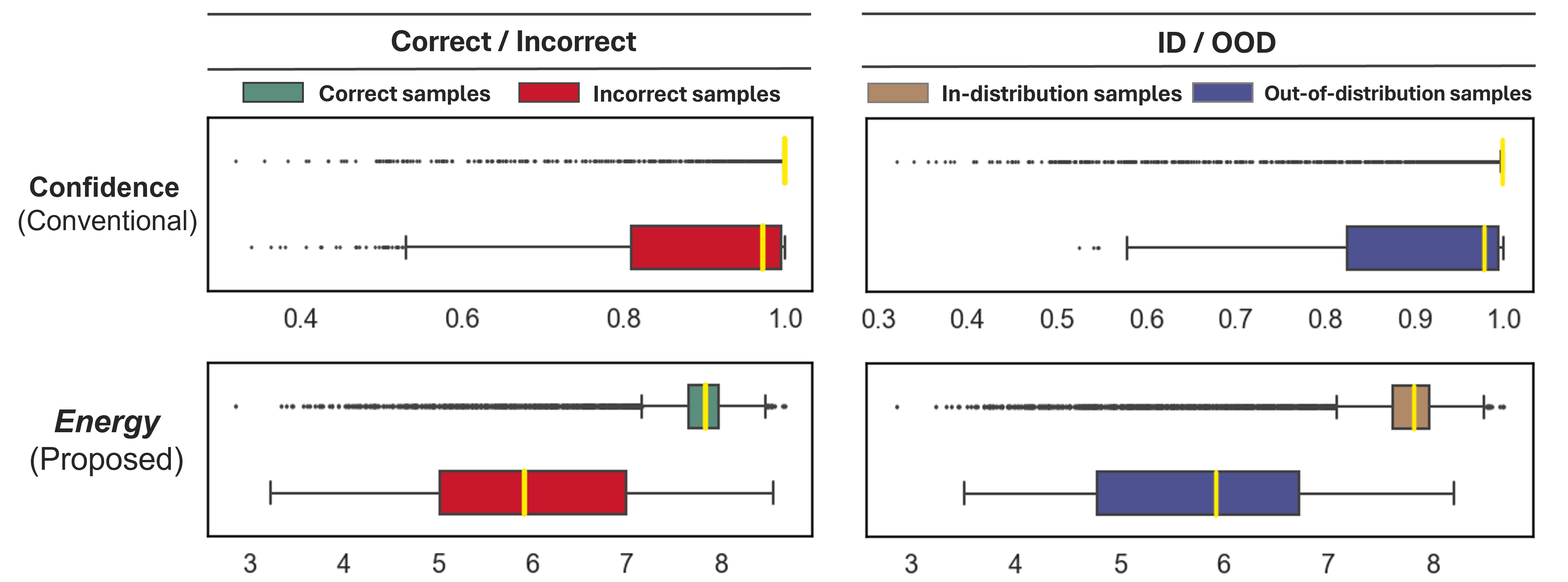}
            \vspace{-4mm}
\caption{\textbf{Conventional softmax confidence scores versus the proposed energy scores.} 
First column: Softmax confidence score (top) and negative energy score (bottom) for correct and incorrect samples. Second column: Softmax confidence score (top) and negative energy score (bottom) for in-distribution (CIFAR10) and OOD samples (SVHN). Our energy scores exhibit greater separability between correct and incorrect predictions, as well as between in-distribution and OOD samples in DenseNet201.}
\label{fig:energyplot}
        \vspace{-7mm}
\end{figure} 

Despite the impressive performance demonstrated by recent AI systems, their deployment should be carefully considered, particularly in safety critical situations (\eg autonomous driving, finance, health care, and medical diagnosis), because these systems cannot consistently ensure accurate predictions. For example, in the field of medical diagnosis, incorrect predictions have the potential to result in catastrophic outcomes. To address this concern, the system must transparently reveal the uncertainty associated with its prediction. Recently, DNNs rely on \textit{confidence} as a way of expressing uncertainty, but they tend to assign higher confidence scores than their actual accuracy \cite{guo2017calibration}. This discrepancy  stems from the inherent inability of DNNs to express appropriate uncertainty during inference. 

To solve this problem, an increasing body of research has focused on refining confidence representations to reflect the uncertainty, a field known as uncertainty calibration. This effort aims to adjust confidence scores to align more closely with accuracy, ultimately improving the reliability of predictions.
As a result of active researches in this field \cite{mukhoti2020calibrating,muller2019does,guo2017calibration,rahimi2020intra,gupta2020calibration,zhang2020mix}, the discrepancy between confidence and accuracy has been significantly mitigated. However, these methods mainly deal with samples drawn from the same distribution on which DNNs were trained (\ie in-distribution), often overlooking distribution shift scenarios. Consequently, they face difficulties in achieving calibration effects when confronted with distribution shift scenarios. 
When considering real-world deployment, calibration methods should demonstrate robustness in handling samples from unknown distributions. In this context, this issue has been considered by Tomani et al. \cite{tomani2021post} within the domain of post-hoc calibration. This method exhibits relatively effective performance in out-of-distribution (OOD) scenarios. However, conversely, it suffered from miscalibration in-distribution (ID) scenarios, exhibiting even greater miscalibration compared to the pre-calibration state.

In this paper, we propose a novel energy-based calibration method that exhibits robustness across the spectrum, from ID to OOD scenarios, including various distribution shift scenarios. We address uncertainty calibration in the context of multi-class classification, particularly in a post-hoc manner. The proposed method utilizes the \textit{energy score} to adeptly capture the uncertainty in DNNs. Previous studies \cite{liu2020energy} have shown that the energy model produces scores that are discriminative between ID and OOD samples.
Beyond this work, we demonstrate that the energy function can induce distinctive scores not only between ID and OOD samples but also between correct and incorrect samples, which can be effectively utilized for uncertainty calibration. Before delving into the mathematical derivation for this in Section \ref{sec:proposed method}, we provide an intuitive overview of the proposed energy score in Fig.\ref{fig:energyplot}. The energy score exhibits a superior ability to produce distinctive scores between ID and OOD samples, as well as between correct and incorrect samples.
This implies that the energy score can more accurately represent the uncertainty of DNNs compared to the confidence score.
Motivated by this observation, we utilize the energy score as an uncertainty estimator for each prediction, which inspires us to propose an instance-wise robust calibration method that adjusts the calibration factor accordingly.
In Section \ref{sec:Experiment}, we demonstrate that the proposed method achieves remarkable robust calibration performance across various baseline DNN models in the wild datasets. This includes scenarios involving OOD samples with semantic shift and varying degrees of covariate shift, as well as ID scenarios.
To sum up, our main contributions are as follows:
\begin{itemize}
    \item We demonstrate the effectiveness of utilizing energy scores for uncertainty calibration through both mathematical derivation and empirical validation.
    \item We introduce a novel post-hoc calibration method that utilizes the energy score to adaptively capture the uncertainty of predictions in DNNs for each individual input.
    \item We illustrate that the proposed calibration method shows robustness in the wild datasets across a wide range of distribution shifts such as covariate and semantic shifts, as well as in complete ID.
\end{itemize}

%% file: sec/2_related_work.tex
\section{Related work}
\label{sec:related work}

\subsection{Post-hoc Calibration} 
Confidence calibration can be divided into two categories. The first category is known as \textit{training-time calibration} \cite{muller2019does, mukhoti2020calibrating, thulasidasan2019mixup,zhong2021improving, liu2022devil}, such as focal loss \cite{mukhoti2020calibrating} and label smoothing \cite{muller2019does}. These methods train DNNs to exhibit calibrated behavior during training. The second category is referred to as \textit{post-hoc calibration} \cite{guo2017calibration, zhang2020mix, rahimi2020intra, wenger2020non, hekler2023test}. In this approach, pre-trained neural networks are utilized along with hold-out validation datasets to learn calibration mappings in a post-hoc manner.

Post-hoc calibration can be further categorized into non-parametric and parametric approaches. Non-parametric approaches include Histogram Binning (HB) \cite{zadrozny2001obtaining} and Isotonic Regression (IR) \cite{zadrozny2002transforming}. HB divided predicted probabilities into multiple intervals, each associated with representative confidences. IR utilized isotonic regression with uncalibrated confidences as the x-axis and the expected accuracy values as the y-axis. BBQ \cite{naeini2015obtaining}, non-parametric extension of HB, incorporated Bayesian model averaging to enhance calibration.
On the other hand, the most common parametric calibration method is Temperature Scaling (TS) \cite{guo2017calibration}. As the temperature increases, the distribution of logits becomes more uniformly distributed, resulting in a decrease in the confidence score associated with the predicted label.
TS has a significant advantage in terms of accuracy preservation, as it maintains the originally predicted label with the highest confidence score unchanged.
However, TS exhibits limited expressiveness as it relies on only a single parameter that is fixed on the validation set.

To address this issue, ensemble approaches, which combine both non-parametric and parametric methods, have been proposed. Ensemble TS \cite{zhang2020mix} introduced additional parameters to enhance expressiveness, building upon TS. IRM \cite{zhang2020mix} leveraged the accuracy-preserving property of the parametric approach and the expressiveness of the non-parametric approach, representing a multi-class extension of IR.
Parameterized TS \cite{tomani2022parameterized} employed a similar strategy to address the expressiveness limitation of TS.
In other approaches, Beta calibration \cite{kull2017beta} was extended to Dirichlet calibration \cite{kull2019beyond} and Spline calibration \cite{gupta2020calibration} utilized spline fitting to approximate the empirical cumulative distribution.

We employ a post-hoc calibration approach, however, our method differs from this body of work.
While these methods only addressed scenarios where test samples are drawn from the same distribution on which DNNs were trained, our approach can handle more diverse situations.
It ensures effective calibration for ID samples while maintaining the original classifier accuracy.

\subsection{Beyond In-distribution Calibration}
Conventional research has mainly focused on investigating post-hoc uncertainty calibration methods. However, these methods often overlook scenarios involving distribution shifts.
Due to their dependence on a fixed calibration map optimized for ID validation sets, they struggle to effectively handle unknown test samples.

Recently, the importance of the robustness of post-hoc calibration method across various distribution shift scenarios has been emphasized \cite{tomani2021post, tomani2023beyond}. In \cite{tomani2021post}, various degrees of Gaussian perturbations were injected into the ID validation dataset. The parameters of the calibration method were then adjusted using the perturbed validation data, resulting in enhanced robustness against shifted distributions. However, this method tends to achieve notable performance only within scenarios where a certain degree of distribution shift is present. Moreover, in a complete ID scenario, it exhibits even worse calibration compared to the pre-calibration state.
To solve this problem, DAC \cite{tomani2023beyond} has been proposed as a pre-processing step before employing the existing post-hoc calibration methods. Notably, it leveraged additional output information for uncertainty estimation and enhanced the calibration performance in distribution shift scenarios.

Similar to these methods, we focus on a broad range of scenarios, ranging from ID to various OOD scenarios.
However, unlike these methods, our approach can facilitate robust calibration without requiring any additional DNN's layer information except for the last layer. In experiments, we demonstrate that our proposed method performs comparably to, and in some cases even surpasses, other state-of-the-art methods that utilize DAC as a preprocessing step.

%% file: sec/3_problem_setup.tex
\section{Problem Setup}
\label{sec:Problem Setup}

In this section, we establish preliminaries for uncertainty calibration. We define key notations in the context of multi-class classification and present a representative calibration metric derived from the concept of perfect calibration. Additionally, we address calibration in OOD scenarios.

\subsection{Notation}

Let $\mathbf{x} \in \mathcal{X}$ and $y \in \mathcal{Y}=\{1, ..., K\}$ be random variables that denote $d$-dimensional inputs and labels, respectively, in multi-class classification tasks with $K$ classes.
These random variables follow a joint distribution $\pi(\mathbf{x},y) = \pi(y | \mathbf{x})\pi(\mathbf{x})$. The dataset  $\mathcal{D} = \{(\mathbf{x}_{n}, y_{n})\}^{N}_{n=1}$ consists of the $N$ number of i.i.d. samples drawn from $\pi(\mathbf{x},y)$. Let $f$ be a pre-trained neural network and $f(\mathbf{x}) = (\hat{y}, \mathbf{z})$ be the output of the neural network, where $\hat{y}$ is a predicted label and $\mathbf{z}$ is an original non-probabilistic output of the network, referred to as the logit. The logit $\mathbf{z}$ is converted into a probabilistic value $\hat{p}$ using the softmax function $\sigma_{SM}$. Thus, $\hat{p}$ represents a confidence score associated with the predicted label $\hat{y}$. To summarize, the output of the neural classifier $f$, $\hat{p}$ and $\hat{y}$, can be obtained as follows:
\[
\hat{p} = \operatorname*{max}_k\sigma_{SM}(\mathbf{z}),~\text{and}\quad \hat{y} = \operatorname*{argmax}_k \sigma_{SM}(\mathbf{z})\quad\text{for}\quad k \in \{1, ..., K\}. 
\]

\subsection{Calibration Metric} 
Perfect calibration is achieved when the predicted probability (\ie confidence) matches with the actual likelihood of a correct prediction. 
If a neural network predicts a label as $y$ with confidence $p$, the actual likelihood of the prediction should ideally be $p$.
Thus, the perfect calibration in multi-class classification can be represented as follows:
\begin{equation}\label{eqn:perfect calibration}
  \mathbb{P}(\hat{y} = y | \hat{p} = p) = p, \quad \forall p \in [0,1].  
\end{equation}
The goal of uncertainty calibration is to minimize the gap between the ground-truth likelihood and the predicted confidence by calibrating the confidence value.

Using this definition of perfect calibration, the calibration error can be computed by modifying \eqref{eqn:perfect calibration}:
\begin{equation}\label{eqn:perf error}
    \mathbb{E}_{\hat{p}}\left[ \left| \mathbb{P}(\hat{y} = y | \hat{p} = p) - p \right| \right].
\end{equation}
Subsequently, the expected calibration error (ECE) \cite{naeini2015obtaining} empirically approximates the calibration error in \eqref{eqn:perf error} using a binning technique. By discretizing the confidence interval into $M$ equally sized bins, \ie $\{B_{m}\}^{M}_{m=1}$, the ECE calculates a weighted average of differences between accuracy $\text{acc}(\cdot)$ and confidence $\text{conf}(\cdot)$ in each bin. With $N$ samples, the ECE is defined as follows:
\begin{equation}\label{eqn:ece}
    \text{ECE} = \sum_{m=1}^{M} \cfrac{|B_{m}|}{N} \left| \text{acc}(B_{m}) - \text{conf}(B_{m}) \right|,
\end{equation}
where all ECE values in this paper are calculated with $M=15$ and multiplied by $100$. In addition to ECE, there are other metrics such as Maximum Calibration Error (MCE)\cite{naeini2015obtaining}, which represents the highest error among bins, Static Calibration Error (SCE)\cite{nixon2019measuring} that evaluates calibration errors on a classwise manner, and KDE-ECE\cite{zhang2020mix} that utilizes Kernel Density Estimation (KDE). 
As ECE is the most representative metric, we primarily evaluate the proposed method using ECE, but we also employ other metrics.

\subsection{Calibration in OOD Scenarios}
In general, OOD refers to a distribution that differs from the training distribution \cite{hsu2020generalized, yang2021generalized}. In this paper, the term OOD includes two types of distribution shifts: \textit{covariate shift} and \textit{semantic shift}. Covariate-shifted samples are drawn from a different joint distribution $\pi_{ood}^{cov}(\mathbf{x},y)$ such that $\pi_{ood}^{cov} \neq \pi$. In other words, while the samples may belong to the same class, they are presented in different forms \cite{patel2015visual}. In the case of semantic shift, the data is drawn from $\pi_{ood}^{sem}(\mathbf{x},\bar{y})$, where $\mathcal{Y} \cap \bar{\mathcal{Y}} = \emptyset$, indicating that the data is from classes not present in the training set $\mathcal{D}$ \cite{hsu2020generalized}. Therefore, in semantic shift scenarios, all predictions by a pre-trained classifier may be incorrect, as they may correspond to one of the K in-distribution classes in $\mathcal{Y}$. From the perspective of calibration, in such scenarios, the lower the confidence, the better the calibration.

%% file: sec/4_proposed_method.tex
\section{Proposed Method}
\label{sec:proposed method}
        \vspace{-5mm}
\begin{figure*}[ht!]
\centering
    \includegraphics[width=\textwidth]{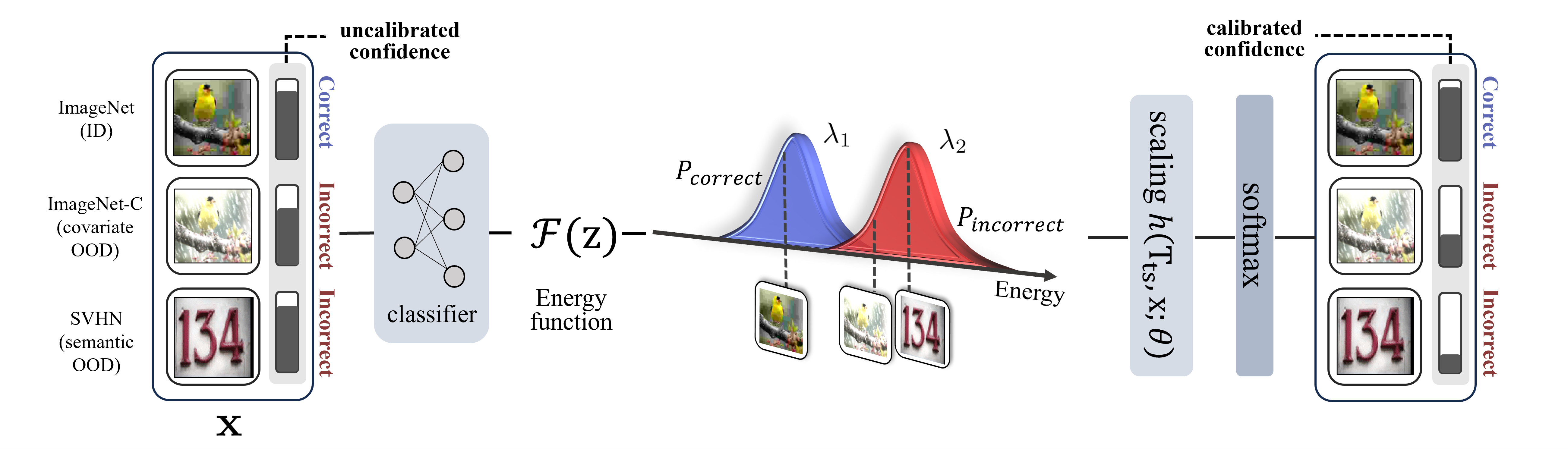}
        \vspace{-7mm}
\caption{\textbf{The overall pipeline for our calibration method.} An input image $\mathbf{x}$ from the wild dataset is fed into a pre-trained classifier, producing the logit $\mathbf{z}$. Subsequently, the free energy $\mathcal{F}$ defined in \eqref{ref:free energy} is calculated for each logit $\mathbf{z}$. Then, two probability density functions (PDFs), \ie $P_{correct}$ and $P_{incorrect}$, are estimated based on the free energy $\mathcal{F}$ of correct and incorrect instances, respectively. These PDFs are utilized to adjust the scaling factors $\lambda_{1}$ and $\lambda_{2}$ in \eqref{ref:lambda}. With the scaling factors, the parameters $\theta_{1}$ and $\theta_{2}$ are then trained through the optimization of the loss function in \eqref{ref:obj}. Using the trained parameters, the calibrated confidence for a test image $\mathbf{x}$ can be calculated by applying scaling in an instance-wise manner as in \eqref{ref:calibrated confidence}.}
\label{fig:pipeline}
        \vspace{-5mm}
\end{figure*} 

To overcome the limitations of conventional calibration methods when distribution shifts exist, we propose a robust calibration method that exhibits calibration improvements across various OOD scenarios.
Previous approaches \cite{tomani2021post} that initially addressed shift scenarios in post-hoc calibration could not properly handle ID inputs. In contrast, our method achieves calibration improvements in both OOD and ID scenarios by adaptively capturing the uncertainty of pre-trained neural networks for each input. To accomplish this, our method leverages the concept of the energy model, which is technically derived from energy-based OOD detection methods \cite{liu2020energy}. Before introducing our calibration method, we lay out the mathematical motivation behind the proposed method by establishing a correlation between our method and the energy model. The overall pipeline for understanding the propose method is illustrated in Fig.\ref{fig:pipeline}.

\subsection{Mathematical Motivation}
The energy function makes scores on ID and OOD more distinguishable than the softmax function \cite{liu2020energy}. As demonstrated in \cite{lecun2006tutorial}, there is a connection between Gibbs distributions (\ie Boltzmann distributions) and softmax functions:
\begin{equation} \label{ref:gibbs distribution ref:softmax function}
    P(y|\mathbf{x}) = \frac{e^{- \beta E(\mathbf{x}, y)}}{\int_{y \in \mathcal{Y}} e^{-\beta E(\mathbf{x}, y)}},~\text{and}~
    \sigma_{SM}(y|\mathbf{x}) = \frac{e^{f_y(\mathbf{x})}}{\sum_{i=1}^{K} e^{f_i(\mathbf{x})}}, 
\end{equation}
where $P(y|\mathbf{x})$ denotes the Gibbs distribution with the energy $E(\mathbf{x},y) : \mathbb{R}^{D} \rightarrow \mathbb{R}$ and $\sigma_{SM}(y|\mathbf{x})$ indicates the softmax function for the $K$-class classifier $f(\mathbf{x}) : \mathbb{R}^{D} \rightarrow \mathbb{R}^{K}$. In \eqref{ref:gibbs distribution ref:softmax function}, $f(\mathbf{x})$ outputs a vector of length $K$ and $f_i(\mathbf{x})$ denotes the $i$-th element of the vector.

By comparing $P(y|\mathbf{x})$ and $\sigma_{SM}(y|\mathbf{x})$ in \eqref{ref:gibbs distribution ref:softmax function}, we can derive the energy as 
\begin{equation}\label{ref:energy}
    E(\mathbf{x}, y) = -f_y(\mathbf{x}), 
\end{equation}    
where the positive constant value $\beta$ is set to $1$. The denominator of $P(y|\mathbf{x})$ in \eqref{ref:gibbs distribution ref:softmax function} is a partition function, transforming each energy value corresponding to $y$ into a probability value within the range of $[0,1]$. In particular, Helmholtz free energy is defined as a log partition function \cite{lecun2006tutorial}. Then, the free energy $\mathcal{F}$ can be represented using the connection between the Gibbs distribution and the softmax function: 
\begin{equation}\label{ref:free energy}
    \mathcal{F}(\mathbf{x}) = - \log \sum_{i=1}^{K} e^{f_i(\mathbf{x})}.
\end{equation}

A mathematical relationship between the energy and the negative log likelihood (NLL) loss has been derived in \cite{lecun2006tutorial}. Based on this, it is demonstrated that the NLL loss inherently decreases the energy for ID samples, while increasing the energy for OOD samples \cite{liu2020energy}.
From these findings, we define the NLL loss, $\mathcal{L}_{NLL}=-\log \sigma_{SM}(y|\mathbf{x})$, as a combination of $E(\mathbf{x}, y)$ in \eqref{ref:energy} and $-\mathcal{F}(\mathbf{x})$ in \eqref{ref:free energy}.  
\begin{equation} \label{ref:NLL}
\begin{split}
    \mathcal{L}_{NLL} =  -\log \frac{e^{f_y(\mathbf{x})}}{\sum_{i=1}^{K} e^{f_i(\mathbf{x})}} =  -f_y(\mathbf{x}) + \log \sum_{i=1}^{K} e^{f_i(\mathbf{x})}
    = E(\mathbf{x}, y) - \mathcal{F}(\mathbf{x}),
\end{split}    
\end{equation}
where the free energy $\mathcal{F}$ can be interpreted as a contrastive term that aggregates the energies for all classes of $i \in \{{1,...,K}\}$.
From the third equation in \eqref{ref:NLL}, we can see that the NLL loss inherently lowers the energy for the correct label $y$ and raises the energy for the other labels.
Additionally, the derivative of $\mathcal{L}_{NLL}$ over the network parameter $\theta$ is calculated as follows.
\begin{equation} \label{ref:derivative NLL}
\begin{split}
    \frac{\partial \mathcal{L}_{NLL}}{\partial \theta} &= \frac{\partial E(\mathbf{x},y)}{\partial \theta} - \frac{\partial \mathcal{F}(\mathbf{x})}{\partial \theta}
    = \frac{\partial E(\mathbf{x},y)}{\partial \theta} - \sum^{K}_{i=1} \frac{\partial E(\mathbf{x},i)}{\partial \theta} \frac{e^{-E(\mathbf{x},i)}}{\sum^{K}_{j=1}e^{-E(\mathbf{x},j)}}
    \\ 
   &= \frac{\partial E(\mathbf{x},y)}{\partial \theta} - \sum^{K}_{i=1} \frac{\partial E(\mathbf{x},i)}{\partial \theta} P(i|\mathbf{x})
   \\
   &= \frac{\partial E(\mathbf{x},y)}{\partial \theta} (1-P(y|\mathbf{x})) - \sum^{K}_{i\neq y} \frac{\partial E(\mathbf{x}, i)}{\partial \theta} P(i|\mathbf{x}), 
\end{split}    
\end{equation}
where the second equality holds by using the Gibbs distribution in \eqref{ref:gibbs distribution ref:softmax function} ($\beta = 1$). 
From the last equation in \eqref{ref:derivative NLL}, we can see that the energy function is weighted by each probability, pushing down the energy for the correct label and pulling up the energy for the incorrect labels.
Liu et al. \cite{liu2020energy} explained that increasing the energy, except for the correct labels, inherently boosts the energy of OOD samples. Following this, the free energy $\mathcal{F}$ can serve as the \textit{energy score}, which can make distinctive values between ID and OOD. This is because it is a smooth approximation of $E$ with the dominance of the ground-truth label $y$ over all other labels. They focused on the distinguishability of the energy score on ID and OOD samples in the OOD detection task, particularly in cases of semantic shift. To further expand this interpretation,  we bring the concept into the perspective of uncertainty calibration.

In our context, we focus on not only the distinguishability between \textit{ID} and \textit{OOD} samples but also between \textit{correct} and \textit{incorrect} predictions. Interpreting the final equation in \eqref{ref:derivative NLL} in a simpler manner, it reveals that the energy score has the capability to differentiate between correct and incorrect samples. This implies that it can make distinctive scores between correct and incorrect predictions not only in ID but also in covariate shifts, as $\mathcal{Y}$ remains consistent. Furthermore, for semantic shift cases, as demonstrated in \cite{liu2020energy}, all predictions are considered incorrect, indicating distinguishability since there is no overlap in labels ( $\mathcal{Y} \cap \mathcal{\bar{Y}} = \emptyset$). Building upon the motivation outlined so far, we introduce the incorporation into post-hoc uncertainty calibration in the next section.

\subsection{Robust Instance-wise Calibration}
Most existing calibration methods are limited by the assumption of the same distribution on which the classifier has been trained. As the parameters are optimized using the consistent distribution of the validation set, these methods lack the adaptiveness to effectively address distribution shift scenarios. To solve this problem, an uncertainty calibration method should possess the capability to capture the uncertainty of the neural network for each individual sample. In this context, the energy score can effectively fulfill this role within the framework of post-hoc calibration. As shown in the motivation and Fig.\ref{fig:energyplot}, it is evident that the energy score is capable of producing distinct values for both cases: ID and OOD samples, as well as correct and incorrect samples.

By facilitating our motivation, we adjust the scaling factor for each input samples to achieve uncertainty calibration. Our method is fundamentally built upon the temperature scaling (TS) technique introduced in \cite{guo2017calibration} to incorporate the advantages of accuracy-preserving property. The proposed scaling factor is defined as follows:
\begin{equation}\label{ref:scaling factor}
    h(T_{ts}, \mathbf{x};\theta) = \underbrace{T_{ts}}_\text{Fixed on validation set} \underbrace{- \lambda_{1}\theta_{1} + \lambda_{2}\theta_{2}}_\text{\, Adaptive for each input},
\end{equation}
where $T_{ts}$ denotes the temperature parameter obtained by the TS technique \cite{guo2017calibration}, which is fixed on the validation set, and $\mathbf{\theta}=\{\mathbf{\theta}_1, \mathbf{\theta}_2\}$ denotes trainable parameters that are optimized using the loss function described in (\ref{ref:obj}). The term $- \lambda_{1}\theta_{1}$ is designed to lower the temperature, whereas $\lambda_{2}\theta_{2}$ is included to raise the temperature. Raising the temperature makes the distribution of the logits to be more uniformly distributed, ultimately reducing confidence in predictions. These adaptive terms adjust the scaling factor depending on whether a given sample is likely to be correctly classified or not.

\begin{algorithm}[t]
\caption{Training Calibration Parameters}\label{alg: train}
\begin{algorithmic}
    \REQUIRE  In-distribution validation data: $\mathcal{D}_{in}$, semantic OOD data $\mathcal{D}_{out}$, Pre-trained $K$-class classifier: $f(\mathbf{x})$, Temperature value from temperature scaling: $T_{ts}$\\
    \STATE 
    \STATE 
    $\mathcal{D} = \mathcal{D}_{in} + \mathcal{D}_{out}$\\
    $(\mathbf{x}, y) \sim \mathcal{D}$\\
    $ \mathbf{z} = f(\mathbf{x})$\\
    $\mathcal{F}(\mathbf{z}) \leftarrow -\texttt{logsumexp}(\mathbf{z})$\\
    \STATE 
    \STATE Estimate Gaussian probability density functions \\
    $P_{1} \leftarrow \mathrm{fit}(\mathcal{F}(\mathbf{z}_{correct}))$\\
    $P_{2} \leftarrow \mathrm{fit}(\mathcal{F}(\mathbf{z}_{incorrect}))$
    \STATE 
    \FOR{$i \in [1, N]$}
    \STATE
    Find $\mathbf{\theta}$ that minimizes the MSE loss\\
    $T = T_{ts} - P_{1}(\mathcal{F}(\mathbf{z}_{i}))\theta_{1} + P_{2}(\mathcal{F}(\mathbf{z}_{i}))\theta_{2}$\\
    $\mathcal{L} = MSE(\texttt{softmax}(\mathbf{z}_{i}/T), y_{i})$
    \ENDFOR
    \STATE 
    \STATE \textbf{return} $\mathbf{\theta}$, $P_{1}$, $P_{2}$\\
    
\end{algorithmic} 
\end{algorithm}
We obtain $\lambda_{1}$ and $\lambda_{2}$ in \eqref{ref:scaling factor}, as follows:
\begin{equation} \label{ref:lambda}
    \lambda_{1} = P_{correct}(\mathcal{F}(\mathbf{x})),~\text{and}~
    \lambda_{2} = P_{incorrect}(\mathcal{F}(\mathbf{x})),
\end{equation}
where $\mathcal{F}$ denotes the energy score function in \eqref{ref:free energy}, and $P_{correct}$ and $P_{incorrect}$ denote the probability density functions of $\mathcal{F}(\mathbf{x})$ fitted to Gaussian distributions for correct and incorrect samples, respectively. By leveraging the energy scores that characterize a specific classifier on a given dataset, we can establish a distribution for these energy scores and subsequently calculate the corresponding probability density function. The energy scores corresponding to correct instances are utilized to construct the distribution of $P_{correct}$, whereas the energy scores associated with incorrect instances are used to form the distribution of $P_{incorrect}$. We utilize an ID validation set and a semantic OOD dataset to make the aforementioned probability density functions, in which the predictions made using semantic OOD data are categorized as incorrect instances, since all predictions correspond to one of the K in-distribution classes.

\begin{algorithm}[t]
\caption{Proposed Instance-wise Calibration Method} \label{alg : apply}
\begin{algorithmic}
    \REQUIRE  Test data : $\mathcal{D}$, Pre-trained $K$-class classifier $f(\mathbf{x})$, Temperature value from temperature scaling: $T_{ts}$, Trained parameters: $\theta$,  PDFs: $P_{1}$, $P_{2}$\\
    \STATE 
    \STATE 
    $(\mathbf{x}, y) \sim \mathcal{D}$\\
    $ \mathbf{z} = f(\mathbf{x})$\\
    $\mathcal{F}(\mathbf{z}) \leftarrow -\texttt{logsumexp}(\mathbf{z})$\\

    \STATE
    $T = T_{ts} - P_{1}(\mathcal{F}(\mathbf{z}))\theta_{1} + P_{2}(\mathcal{F}(\mathbf{z}))\theta_{2}$\\
    $\hat{q}= \max(\texttt{softmax}(\mathbf{z}/T))$
    \STATE 
    \STATE \textbf{return} calibrated confidence $\hat{q}$\\
    
\end{algorithmic} 
\end{algorithm}

By designing the proposed scaling factor in the manner described above, our method can gain the ability to instance-wisely distinguish between incorrect and correct samples. Ensuring the adaptability of our calibration method to each input sample is crucial for capturing the uncertainty of a particular prediction, especially in the presence of distribution shifts. This is the main reason on the robust calibration performance exhibited by our method across various test data, including ID samples and various types of distribution shifted samples (Fig.\ref{fig:lineplot}), which is empirically demonstrated in the experiment section. Then, using our instance-wise scaling with the scaling factor in \eqref{ref:scaling factor}, the calibrated probability of sample $\mathbf{x}$ for $K$ classes, $\hat{p}_{\theta}\in\mathbb{R}^{K}$, can be expressed as follows:
\begin{equation}\label{ref:calibrated probability}
    \hat{p}_{\theta} = \sigma_{SM}(f(\mathbf{x})/h(T_{ts}, \mathbf{x};\theta)),
\end{equation}
where $h(T_{ts}, \mathbf{x};\theta)$ denotes the proposed scaling factor defined in \eqref{ref:scaling factor}.
To train the parameters $\mathbf{\theta}=\{\mathbf{\theta}_1, \mathbf{\theta}_2\}$ in \eqref{ref:calibrated probability}, we design a mean squared error loss function:
\begin{equation}\label{ref:obj}
    \mathcal{L}_{\theta} = \frac{1}{N}\sum_{i}||y^{(i)} - \hat{p}_{\theta}^{(i)}||_2^{2},
\end{equation}
where $y^{(i)}\in \mathbb{R}^{K}$ denotes the one-hot encoded ground-truth label for the $i$-th sample, $\hat{p}_{\theta}^{(i)}\in \mathbb{R}^{K}$ denotes $\hat{p}_{\theta}$ in \eqref{ref:calibrated probability} for the $i$-th sample, and $N$ is the total number of training samples. Using the trained parameter $\theta$, the calibrated confidence for a test sample $\mathbf{x}$ can be calculated in an instance-wise manner:
\begin{equation}\label{ref:calibrated confidence}
    \hat{q} = \max(\sigma_{SM}(f(\mathbf{x})/h(T_{ts}, \mathbf{x};\theta)).
\end{equation}

Algorithm \ref{alg: train} describes the entire procedure for training calibration parameters, while Algorithm \ref{alg : apply} outlines the procedure of implementing the instance-wise calibration method with the trained parameters. 

\vspace{-3mm}

%% file: sec/5_experiment.tex
\section{Experiment}
\label{sec:Experiment}

\subsection{Experimental Settings}

For experiments, we trained classification DNNs including VGGNet  \cite{simonyan2014very}, ResNet \cite{he2016deep}, WideResNet \cite{zagoruyko2016wide}, DenseNet \cite{huang2017densely} and SE-ResNet \cite{hu2018squeeze} on the CIFAR10/CIFAR100 datasets \cite{krizhevsky2009learning}. We employed pre-trained weights implemented in PyTorch for ImageNet-1k \cite{deng2009imagenet}. We utilized two types of datasets for training: one intended for ID scenarios and the other for semantic OOD scenarios.
For tuning $\theta$, we utilized ID validation images (5,000 for CIFAR10/CIFAR100, 12,500 for ImageNet-1k) and semantic OOD samples (100/400/3,500 SVHN \cite{netzer2011reading} or Texture \cite{cimpoi2014describing} images for CIFAR10/CIFAR100/ImageNet-1k, respectively).

To evaluate our method, we employed 10,000/10,000/12,500 images each from CIFAR10/CIFAR100/ImageNet-1k as test ID samples. For covariate OOD test data, we utilized the corrupted dataset CIFAR10-C, CIFAR100-C, and ImageNet-C in \cite{hendrycks2019robustness}. These corrupted datasets contain five severity levels for 19 corruption types (\eg blur, contrast, and frost). We utilized 10,000 images for each severity level of corruption types, maintaining the same approach across all datasets. For the test semantic OOD scenarios, we used Textures or SVHN dataset, which were not used during the tuning time.

To demonstrate the effectiveness of our method, we compared it with five baseline post-hoc calibration methods, which are TS \cite{guo2017calibration}, ETS \cite{zhang2020mix}, IRM \cite{zhang2020mix}, IROvA/IROvATS \cite{zhang2020mix}, and SPLINE \cite{gupta2020calibration}. Among these methods, TS, ETS, IRM, SPLINE, and our method are considered as accuracy-preserving methods. In addition, we compared our method with the state-of-the-art post-hoc calibration method, DAC, proposed in \cite{tomani2023beyond}. Please note that additional experimental results based on the type of semantic OOD dataset can be found in the \texttt{supplementary materials}.

\vspace{-3mm}
\subsection{Ablation Study on Energy Score}
\vspace{-8mm}
\begin{figure}[h!]
\centering
    \includegraphics[width=0.6\columnwidth]{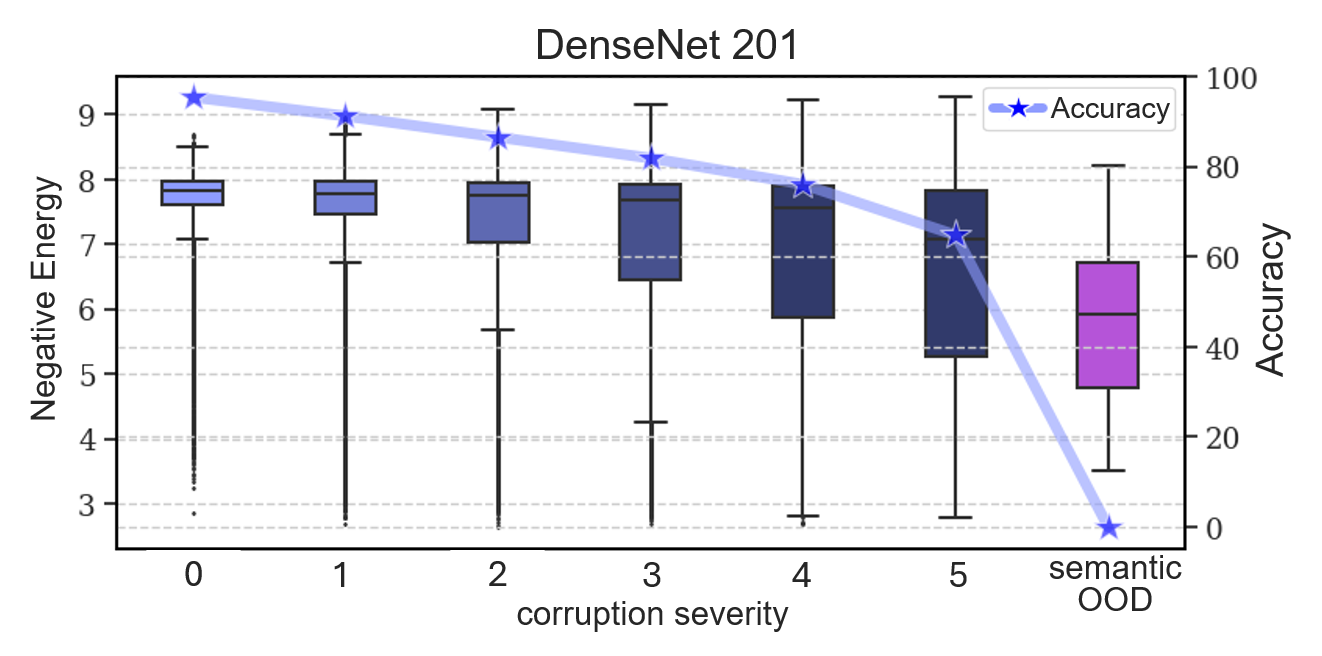}
    \vspace{-3mm}
\caption{\textbf{Negative energy scores and accuracy from ID to semantic OOD data.}
The x-axis shows the severity levels of corruptions, while the left y-axis shows the negative energy scores (box plot) and the right y-axis shows accuracy (line plot). Severity level 0 (light blue) indicates the negative energy score for ID data (CIFAR10), levels 1-5 (deepening shades of blue) represent increasing degrees of corruption (CIFAR10-C), and the negative energy score for semantic OOD data (SVHN) is depicted in purple.} 
\vspace{-5mm}
\label{fig:energy boxplot}
\end{figure}

We conducted an ablation study on the energy score in \eqref{ref:free energy}. To demonstrate the capability of our method in capturing network uncertainty, we evaluated the energy score on various samples from complete ID dataset (CIFAR10), corrupted dataset (CIFAR10-C) and semantic OOD dataset (SVHN). As shown in Fig.\ref{fig:energy boxplot}, the energy scores tend to decrease with higher variances as the degree of distribution shift increases. Since the tendency for energy scores to decrease follows the trend of diminishing accuracy, it implies that the energy scores can indeed efficiently capture the uncertainty of DNNs.

\subsection{Comparison with Baseline Methods}
   \vspace{-5mm}

\begin{table*}[h!]
\centering
\caption{\textbf{Averaged ECE over all severity levels of corruption.} The lowest ECE results were highlighted in bold and the parentheses indicates the OOD dataset used for tuning our method. It can be observed that our method has a notable calibration effect across various DNN architectures and all corruption levels from 0 to 5.}
\label{table:overall ECE}
    \vspace{-3mm}
\centering
\setlength{\tabcolsep}{4pt}
\begin{adjustbox}{width=\textwidth}
\renewcommand{\arraystretch}{0.95}
\begin{tabular}{@{~}l|l|c|ccccccc@{~}}
 \toprule
 \multicolumn{10}{c}{ECE ($\downarrow$)}\\ \toprule
 ID (OOD)&  Network        & Uncalibrated & TS    & ETS   & IRM   & IROvA & IROvATS & SPLINE & \cellcolor{gray!10}Ours          \\
\midrule
\multirow{4}{*}{\begin{tabular}[l]{@{}l@{}}CIFAR10\\ (SVHN)\end{tabular}}        & DenseNet201 & 12.90 & 9.30  & 8.95 & 5.78 & 6.76  & 6.05 & 7.40 & \cellcolor{gray!10}\textbf{4.47} \\
 & VGG19        & 16.17 & 10.59 & 10.34 & 7.18  & 7.92  & 7.33    & 7.76  & \cellcolor{gray!10}\textbf{5.78} \\
 & WideResNet40 & 12.49 & 8.57  & 8.33  & 5.72  & 7.80  & 7.35    & 7.30  & \cellcolor{gray!10}\textbf{4.71} \\
 & SE-ResNet50  & 14.80 & 10.17 & 9.94  & 8.13  & 8.51  & 8.34    & 9.09  & \cellcolor{gray!10}\textbf{6.28} \\
 \midrule
\multirow{4}{*}{\begin{tabular}[l]{@{}l@{}}CIFAR100\\ (SVHN)\end{tabular}}       & DenseNet201 & 22.79 & 11.21 & 9.73 & 8.14 & 10.88 & 8.02 & 7.52 & \cellcolor{gray!10}\textbf{5.94} \\
 & VGG19        & 32.49 & 12.54 & 11.99 & 8.65  & 11.42 & 8.55    & 8.23  & \cellcolor{gray!10}\textbf{7.35} \\
 & WideResNet40 & 23.83 & 13.32 & 11.17 & 10.50 & 13.88 & 9.69    & 9.90  & \cellcolor{gray!10}\textbf{8.74} \\
 & SE-ResNet101 & 30.27 & 15.24 & 12.89 & 10.17 & 13.59 & 10.67   & 10.46 & \cellcolor{gray!10}\textbf{8.95} \\
 \midrule
\multirow{2}{*}{\begin{tabular}[l]{@{}l@{}}ImageNet-1k\\ (Texture)\end{tabular}} & DenseNet201   & 6.99  & 5.49  & 5.49 & 6.37 & 10.67 & 8.95 & 6.44 & \cellcolor{gray!10}\textbf{4.92} \\
 & ResNet18  & 7.32  & 5.57  & 5.57  & 6.77  & 9.58 & 8.06    & 6.83  & \cellcolor{gray!10}\textbf{5.31}\\ \bottomrule
 
\end{tabular}
  \end{adjustbox}
   \vspace{-14mm}
\end{table*}

\begin{figure}[h!]
\centering
    \includegraphics[width=1.0\columnwidth]{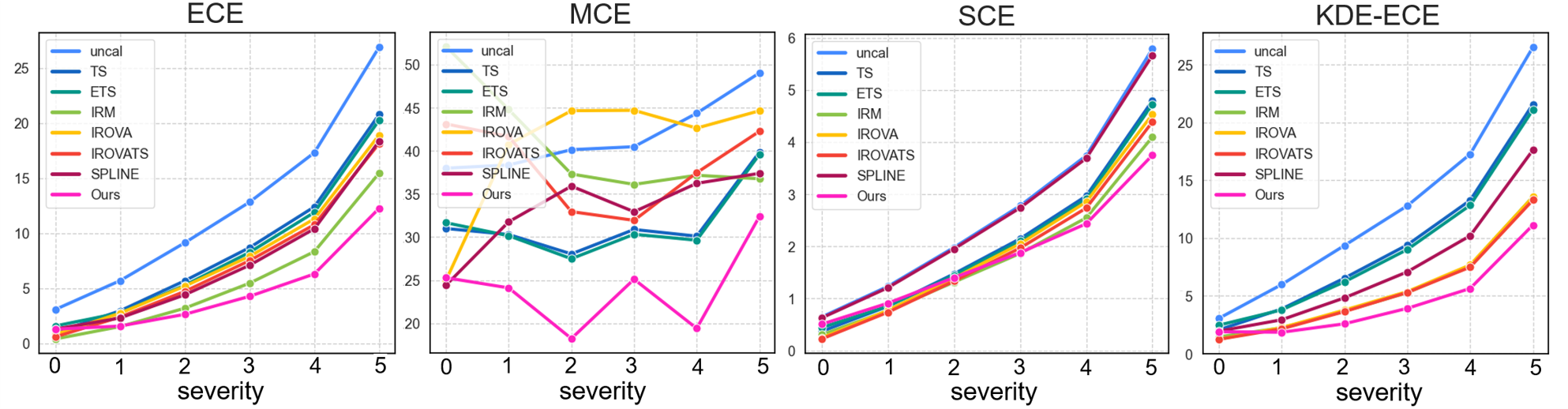}
    \vspace{-7mm}
    \caption{{\textbf{Calibration errors across different levels of corruption severity.}} We can observe that the our method also benefits from the calibration effect in complete ID (severity level 0), while outperforming other approaches in the remaining severity levels. We used WideResNet40 trained on CIFAR10.}
\label{fig:lineplot} 
    \vspace{-6mm}
\end{figure}

We compared our method with the aforementioned baseline methods by evaluating them across various datasets and backbone architectures in terms of ECE. Because our method emphasizes robust calibration performance on diverse datasets, we comprehensively conducted experiments on a variety of distribution shift scenarios, spanning from complete in-distribution to heavily corrupted scenarios. For this purpose, we employed a corrupted dataset comprising severity levels ranging from 1 to 5, with complete ID test data added as severity level 0.

Table \ref{table:overall ECE} shows the averaged ECE across all severity levels. Our method outperforms other baseline methods for various backbone networks and datasets. Furthermore, as shown in Fig.\ref{fig:lineplot}, our method surpassed other approaches in most individual severity levels. It is noteworthy that our method shows consistent performance not only in scenarios involving corruptions but also in cases of complete ID scenarios. This is in contrast to \cite{tomani2021post}, which exhibited greater miscalibration even than the uncalibrated one in the context of complete in-distribution data. 
\begin{figure*}[h!]
\centering
    \includegraphics[width=\textwidth]{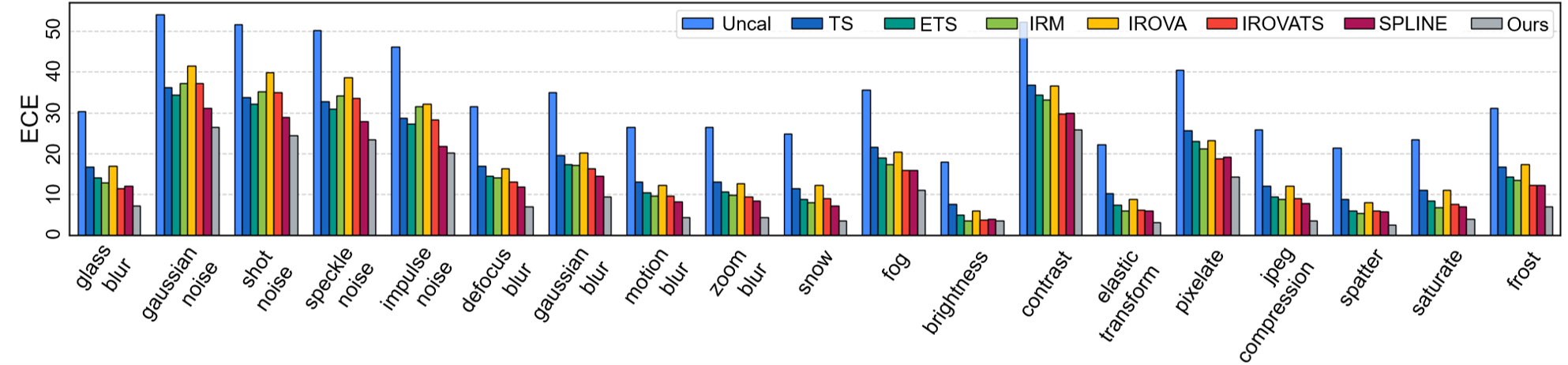}
        \vspace{-7mm}
    \caption{\textbf{Expected Calibration Error (ECE) for various corruption types at severity level 5.} It is evident that our proposed method exhibits superior performance across various types of corruption when compared to other methods, using DenseNet201 trained on the CIFAR100 dataset. }
\label{fig:corruption type}
\vspace{-4mm}
\end{figure*}
Fig.\ref{fig:corruption type} shows a comparison of ECE by corruption type, similar to \cite{tomani2021post, hekler2023test}. Our method demonstrates robust calibration across various corruption types.
Additional results on diverse calibration metrics \cite{nixon2019measuring,zhang2020mix} and transformer-based models \cite{dosovitskiy2020image,liu2021swin} are available in the \texttt{supplementary materials}.
\vspace{-4mm}

\subsection{Exploring Synergies with Applicable State-of-the-art Method}
We analyzed the results of applying an applicable post-hoc calibration method to our proposed method. DAC~\cite{tomani2023beyond}, similar to our objective, aims for robust calibration performance even in OOD scenarios. Unlike most methods, including ours, which use only the output of the last layer of the DNN, DAC additionally utilized the output of other layers.
Since DAC is designed to be used alongside post-hoc calibration methods, we applied it to our proposed method. We followed the DAC's layer selection method proposed by Tomani et al. \cite{tomani2023beyond}. 
Table \ref{table:DAC_all} shows the averaged ECE for each corrupted dataset. We compared our method with ETS+DAC and SPLINE+DAC, both of which primarily achieved state-of-the-art in \cite{tomani2023beyond}. While our method showed good synergy with DAC, it alone achieved the best performance even without DAC. Notably, our approach can attain these results without needing the additional output information from each classifier layer used by DAC.

\vspace{-4mm}
\begin{table}[h]
\caption
{\textbf{ECE comparison on state-of-the-art methods.} The best results were highlighted in bold, while the second best results were emphasized using underline. Our method exhibits comparable performance on its own, even without appending DAC. VGG19 was used for CIFARs, and DenseNet201 for ImageNet.}
\vspace{-3mm}
    \centering
    \setlength{\tabcolsep}{4pt}
    \begin{adjustbox}{width=0.9\textwidth}
    \begin{tabular}{l|c|cc|cc|cc}
        \toprule
        \multirow{2}{*}{Dataset} & \multirow{2}{*}{Uncal.} & \multicolumn{2}{c|}{ETS} & \multicolumn{2}{c|}{SPLINE} & \multicolumn{2}{c}{Ours} \\
        \cmidrule(lr){3-8}
         &  & w/o DAC & w/ DAC & w/o DAC & w/ DAC & w/o DAC & w/ DAC \\
        \midrule 

        CIFAR10-C    & 15.49 & 12.33 & 8.59 & 15.12 & 8.45 & \textbf{5.98} & \underline{7.81} \\
        CIFAR100-C   & 31.83 & 22.21 & 11.97 & 27.97 & 12.28 & \underline{11.92} & \textbf{8.99} \\
        ImageNet-C   & 6.84 & 7.34 & \underline{5.07} & 8.05 & 5.96 & \textbf{4.96} & 5.72 \\
        \bottomrule
    \end{tabular}
    \end{adjustbox}
    \label{table:DAC_all}
\vspace{-9mm}
\end{table}

\subsection{Evaluating Robustness on Semantic OOD}
\begin{figure}[h!]
\centering
    \includegraphics[width=1.0\columnwidth]{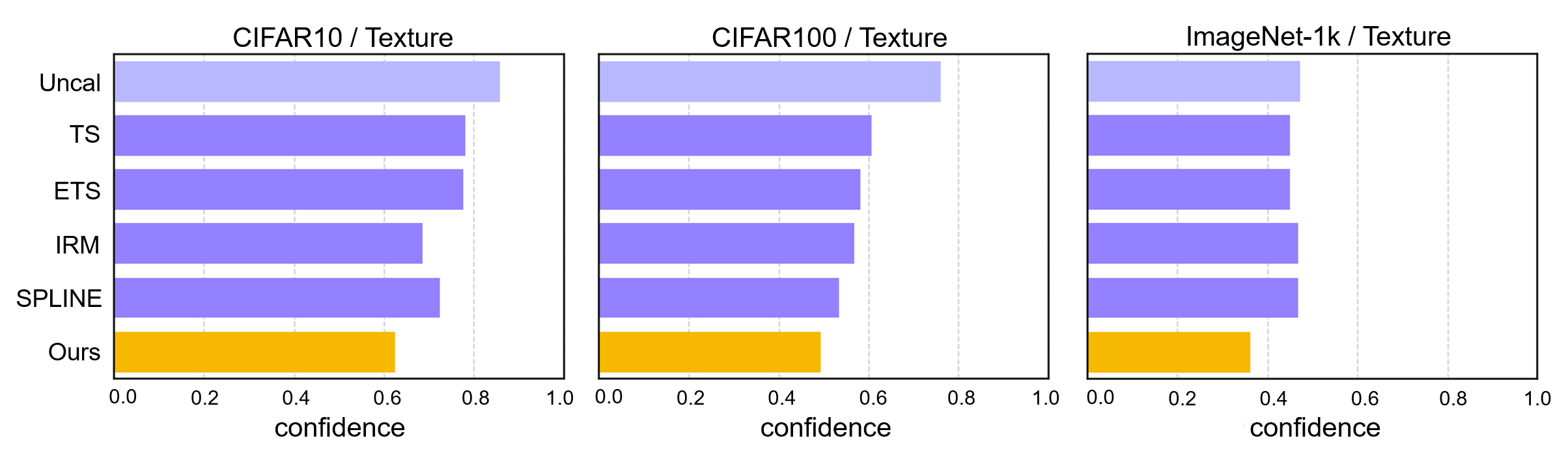}
        \vspace{-9mm}
\caption{\textbf{Calibrated confidence scores for semantic OOD samples.} Each bar denotes the confidence score on semantic OOD samples. The shorter bar represents better-calibrated confidence because all predictions for semantic OOD samples are incorrect; thus, the lower confidence score, the better. For this experiment, we used CIFARs and ImageNet-1k as ID data, while Texture served as the OOD test data.}
\label{fig:oof_confidence}
      \vspace{-1mm}
\end{figure} 

\begin{table}[h]
\caption{\textbf{OOD detection comparison.} In (a) and (b), $[\cdot/\cdot]$ represents the ID/OOD data. In particular, we obtained the results using DenseNet201 trained on CIFAR100 and ImageNet-1k as ID data, while Texture served as the OOD test data.}\label{table:ood_detection}
        \vspace{-3mm}
\centering
\renewcommand{\arraystretch}{0.9}
\begin{minipage}{0.48\textwidth}
\vspace{1mm}
     \centering
     \text{(a) [CIFAR100 / Texture]}
   \end{minipage}
\begin{minipage}{0.48\textwidth}
\vspace{1mm}
     \centering
     \text{(b) [ImageNet-1k / Texture] }
   \end{minipage}
\begin{minipage}{0.48\linewidth}
\centering
    \begin{tabular}{L{1.5cm}C{1.2cm}C{1.2cm}C{1.2cm}}
    \toprule
          & AU-ROC($\uparrow$) & AUPR-IN($\uparrow$) & AUPR-OUT($\uparrow$) \\ \hline
    Uncal.  & 74.21 & 70.62   & 71.25    \\ \hline
    TS     & 74.84 & \textbf{70.89}   & 72.27    \\
    ETS    & 74.86 & \textbf{70.89}   & 72.27    \\
    IRM    & 71.73 & 69.57   & 72.16    \\
    SPLINE & 73.76 & 70.27   & 70.98    \\
    \cellcolor{gray!10}Ours   & \cellcolor{gray!10}\textbf{75.03} & \cellcolor{gray!10}\textbf{70.89}   & \cellcolor{gray!10}\textbf{72.37}    \\ \bottomrule
    \end{tabular}
\end{minipage}
\begin{minipage}{0.48\linewidth}
\centering
    \begin{tabular}{L{1.5cm}C{1.2cm}C{1.2cm}C{1.2cm}}
    \toprule
           & AU-ROC($\uparrow$) & AUPR-IN($\uparrow$) & AUPR-OUT($\uparrow$) \\ \hline
    Uncal.  & 81.43 & 83.50   & 79.17    \\ \hline
    TS     & 81.68 & 83.68   & 79.50    \\
    ETS    & 81.68 & 83.68   & 79.50   \\
    IRM    & 80.25 & \textbf{84.23}  & 79.81    \\
    SPLINE & 81.43 & 83.52   & 79.17    \\
    \cellcolor{gray!10}Ours   & \cellcolor{gray!10}\textbf{82.25} & \cellcolor{gray!10} 83.82   & \cellcolor{gray!10}\textbf{80.14}    \\ \bottomrule
    \end{tabular}
\end{minipage}
\vspace{-5mm}
\end{table}

We measured the calibrated confidence scores for semantic OOD test samples, which are different from OOD train samples used to adjust our calibration parameters. For this experiment, we utilized DenseNet201. Fig.\ref{fig:oof_confidence} demonstrates that our method produces the lowest confidence scores for OOD samples, because all predictions for OOD samples are incorrect.

Furthermore, we conducted an experiment to investigate the potential extension into OOD detection. To accomplish this, we utilized key evaluation metrics commonly employed in OOD detection, such as AUROC, AUPR-in, and AUPR-out. Our method demonstrates superior performance compared to other approaches in most cases, as observed in Table \ref{table:ood_detection}, thus affirming the potential for extension into OOD detection with our proposed approach.

%% file: sec/6_conclusion.tex
\section{Conclusion}
\label{sec:Conclusion}
In this paper, we addressed the limitations of existing post-hoc calibration methods on the wild datasets, including in-distribution, covariate shift, and semantic shift. 
Conventional methods could not consider all these scenarios in achieving robust calibration. To solve this problem, we introduced a novel instance-wise calibration method using the energy score. Our method adaptively captured uncertainty for each instance by leveraging the energy score. Through experiments conducted across various networks and datasets, we demonstrated that our method outperforms existing calibration methods in scenarios involving various types of distribution shifts, while consistently maintaining calibration effect in the complete in-distribution dataset. 

As the reliability of AI in safety-critical situations becomes increasingly important, we believe that our method can contribute to the safer deployment of AI systems in real-world scenarios. By offering a promising direction with our method, we hope to inspire future research efforts for enhancing trustworthy AI.

\subsubsection{Acknowledgements.}
This work was partly supported by the National Research Foundation of Korea
(NRF) grant funded by the Korea government (NRF2020R1C1C1004907) and partly supported by Institute of Information \& communications Technology Planning \& Evaluation (IITP) grant funded by the Korea government(MSIT) (RS-2022-00143911, AI Excellence Global Innovative Leader Education Program and 2021-0-01341,
Artificial Intelligence Graduate School Program (Chung-Ang university)).

%% file: sec/7_supp.tex
\appendix

   \begin{center}
      \Large\textbf{Supplementary Material \\ for Uncertainty Calibration with Energy Based Instance-wise Scaling in the Wild Dataset}\\
   \end{center}
\section{Evaluation on Additional Calibration Metrics}
\label{sec:metrics}

\begin{table*}[ht!]
\vspace{-4mm}
\centering
\caption{\textbf{ Averaged KDE-ECE over all severity levels of corruption. } The lowest KDE-ECE results were highlighted in bold and the parentheses indicates the OOD dataset used for tuning our method, same as in the main text. It can be also observed that our method has a noticeable performance across various DNN architectures and corrupted datasets, when evaluated using KDE-ECE.}
\label{table:overall KDE}
\centering
\setlength{\tabcolsep}{4pt}
\begin{adjustbox}{width=\textwidth}
\renewcommand{\arraystretch}{1.0}
\begin{tabular}{@{~}c|c|c|ccccccc@{~}}
 \toprule
  \multicolumn{10}{c}{KDE-ECE ($\downarrow$)}\\ \hline
   ID(OOD)&  Network        & Uncalibrated & TS    & ETS   & IRM   & IROvA & IROvATS & SPLINE & \cellcolor{gray!10}Ours          \\ \hline
\multirow{4}{*}{\begin{tabular}[c]{@{}c@{}}CIFAR10\\ (SVHN) \end{tabular}}        & DenseNet201 & 12.49 & 9.45  & 9.23 & 5.64 & 5.65  & 5.51 & 7.43 & \cellcolor{gray!10}\textbf{4.50} \\
 & VGG19        & 15.72 & 10.84 & 10.56 & 7.03  & 7.11  & 6.91    & 7.92  & \cellcolor{gray!10}\textbf{5.81} \\
 & WideResNet40 & 12.03 & 8.71  & 8.55  & 5.60  & 6.04  & 5.90    & 7.40  & \cellcolor{gray!10}\textbf{4.72} \\
 & SE-ResNet50  & 14.33 & 10.26 & 10.09  & 7.48  & 7.76  & 7.84    & 9.05  & \cellcolor{gray!10}\textbf{6.31} \\
 \hline
\multirow{4}{*}{\begin{tabular}[c]{@{}c@{}}CIFAR100\\ (SVHN)\end{tabular}}       & DenseNet201 & 22.22 & 11.11 & 9.64 & 10.06 & 10.65 & 9.45 & 8.12 & \cellcolor{gray!10}\textbf{6.07} \\
 & VGG19        & 31.89 & 12.29 & 12.06 & 10.62  & 12.51 & 11.39    & 8.54  & \cellcolor{gray!10}\textbf{7.27} \\
 & WideResNet40 & 23.14 & 13.12 & 11.14 & 12.78 & 14.43 & 12.86    & 10.16  & \cellcolor{gray!10}\textbf{8.69} \\
 & SE-ResNet101 & 29.74 & 15.26 & 13.40 & 13.90 & 14.78 & 13.30   & 10.84 & \cellcolor{gray!10}\textbf{9.19} \\
 \hline
\multirow{2}{*}{\begin{tabular}[c]{@{}c@{}}ImageNet-1k\\ (Texture)\end{tabular}} & DenseNet201   & 6.69  & 5.27  & 5.27 & 5.75 & 13.29 & 13.29 & 6.31 & \cellcolor{gray!10}\textbf{5.11} \\
 & ResNet18  & 7.23  & 5.58  & 5.58  & 6.21  & 14.42 & 14.41    & 6.86  & \cellcolor{gray!10}\textbf{5.29}\\\hline
\end{tabular}
  \end{adjustbox}
\end{table*}

\begin{table*}[ht!]

\caption{\textbf{ Averaged Static Calibration Error (SCE) over all severity levels of corruption. } The lowest SCE results were highlighted in bold and the parentheses indicates the OOD dataset used for tuning our method, same as in the main text. It can be also observed that our method has a noticeable calibration effect across various DNN architectures and corrupted datasets, even when evaluated using SCE.}
\label{table:overall SCE}
\centering
\setlength{\tabcolsep}{4pt}
\begin{adjustbox}{width=\textwidth}
\renewcommand{\arraystretch}{1.0}
\begin{tabular}{@{~}c|c|c|ccccccc@{~}}
 \toprule
  \multicolumn{10}{c}{SCE ($\downarrow$)}\\ \hline
ID(OOD)&  Network & Uncalibrated & TS    & ETS   & IRM   & IROvA & IROvATS & SPLINE & \cellcolor{gray!10}Ours \\ \hline
\multirow{4}{*}{\begin{tabular}[c]{@{}c@{}}CIFAR10\\ (SVHN)\end{tabular}}        & DenseNet201 & 0.56 & 0.42  & 0.41 & 0.43 & 0.46  & 0.43 & 0.61 & \cellcolor{gray!10}\textbf{0.39} \\
 & VGG19        & 3.39 & 2.48 & 2.46 & 2.11  & 2.08  & \textbf{2.06}  & 3.00  & \cellcolor{gray!10}2.07 \\
 & WideResNet40 & 2.70 & 2.10  & 2.08  & \textbf{1.82}  & 1.98  & 1.90    & 2.65  & \cellcolor{gray!10}\textbf{1.82} \\
 & SE-ResNet50  & 3.18 & 2.48 & 2.46  & 2.23  & 2.23  & 2.19    & 3.07  & \cellcolor{gray!10}\textbf{2.18} \\
 \hline
\multirow{4}{*}{\begin{tabular}[c]{@{}c@{}}CIFAR100\\ (SVHN)\end{tabular}}       & DenseNet201 & 0.55 & 0.39 & 0.38 & 0.43 & 0.47 & 0.43 & 0.57 & \cellcolor{gray!10}\textbf{0.37} \\
 & VGG19        & 0.74 & 0.43 & 0.43 & 0.40  & 0.41 & \textbf{0.39}    & 0.64  & \cellcolor{gray!10}\textbf{0.39} \\
 & WideResNet40 & 0.56 & 0.42 & 0.41 & 0.43 & 0.46 & 0.43    & 0.61  & \cellcolor{gray!10}\textbf{0.39} \\
 & SE-ResNet101 & 0.67 & 0.45 & 0.42 & 0.42 & 0.44 & 0.44   & 0.62 & \cellcolor{gray!10}\textbf{0.40} \\
 \hline
\multirow{2}{*}{\begin{tabular}[c]{@{}c@{}}ImageNet-1k\\ (Texture)\end{tabular}} & DenseNet201   & 0.21  & \textbf{0.20}  & \textbf{0.20} & \textbf{0.20} & 0.22 & 0.21 & 0.26 & \cellcolor{gray!10}\textbf{0.20} \\
 & ResNet18  & 0.25  & 0.25  & 0.25  & 0.25  & 0.28 & 0.27 & 0.32  & \cellcolor{gray!10}\textbf{0.24}\\\hline
\end{tabular}
  \end{adjustbox}
\end{table*}

In the main text, we primarily evaluated methods using the Expected Calibration Error (ECE), which remains the most representative metric for assessing calibration. The purpose of this section is to present results evaluated using other calibration metrics besides the ECE introduced in the main text. To this end, we evaluated methods using the Kernel Density Estimation based ECE (KDE-ECE) \cite{zhang2020mix} and the Static Calibration Error (SCE)\cite{nixon2019measuring}. Both metrics are lower when calibration is better. Note that the SCE is conceptually identical to the Class-wise Calibration Error \cite{kull2019beyond}. Additionally, given the characteristic of the spline method, which calibrates confidence for the top label only, it is advisable to note that when evaluating spline using the class-wise SCE, there may be more penalty due to its design of the method. Results for KDE-ECE and SCE are respectively shown in Tables \ref{table:overall KDE} and \ref{table:overall SCE}. Both results are based on the same in-distribution (ID), out-of-distribution (OOD) settings, and likewise, the same networks as utilized in the main text. In most results, it can be observed that our method still shows superior performance with both KDE-ECE and SCE.

\section{Evaluation on Transformer-based Models}
\label{sec:vit}
\vspace{-5mm}

\begin{figure}[ht!]
\centering
    \includegraphics[width=0.6\columnwidth]{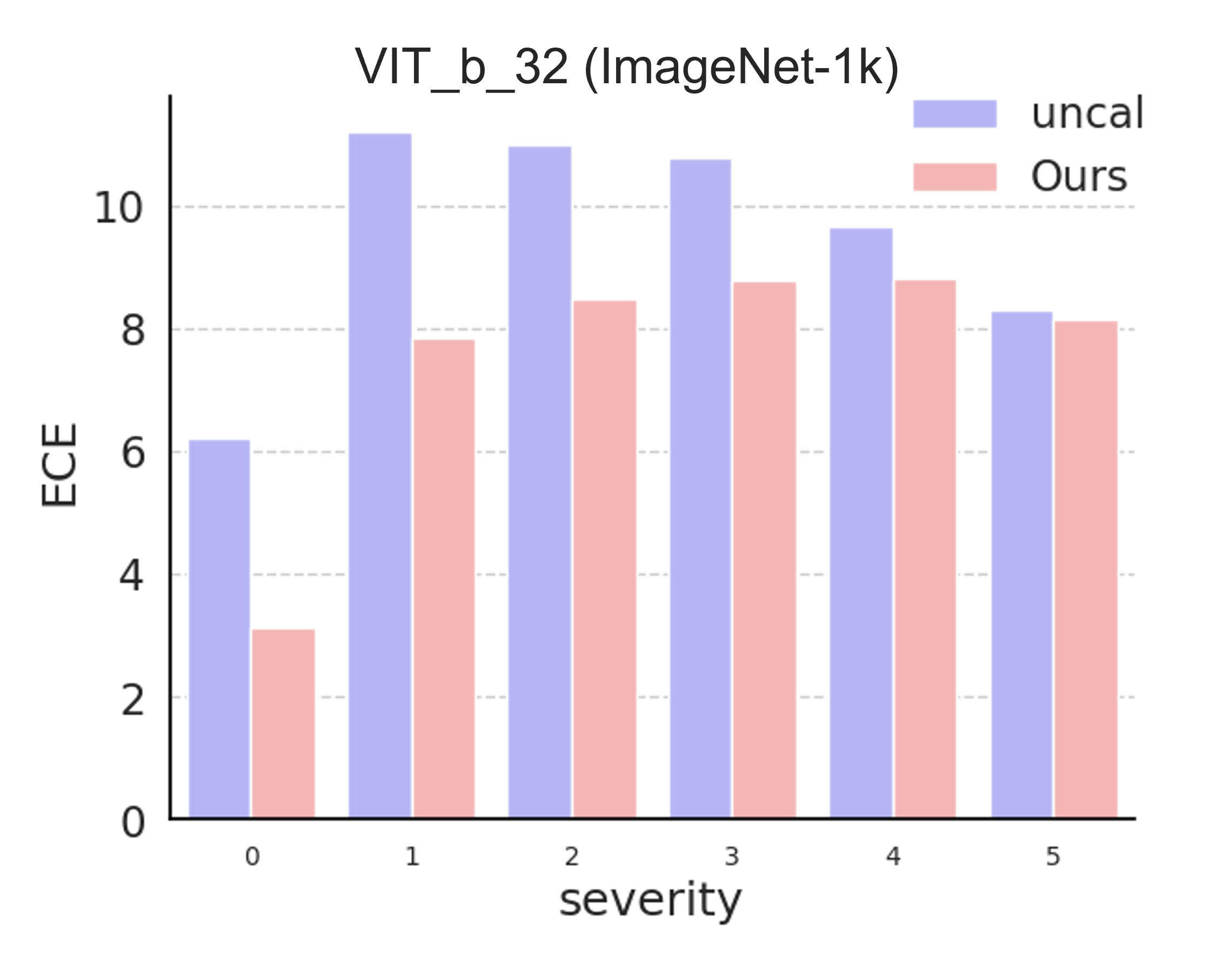}
\caption{\textbf{Expected Calibration Error (ECE) across different levels of corruption severity.} At all levels of corruption severity, the proposed method consistently demonstrates calibration effectiveness in transformer-based vision models. }
\label{fig:vit}
\end{figure}

\begin{table}[ht!]

\centering
\caption{\textbf{Calibration effect across diverse transformer-based models.} Results on averaged ECE and KDE-ECE over all severity levels of corruption. It demonstrates calibration effect of proposed method across diverse transformer-based models.}
\label{table:vit}
\setlength{\tabcolsep}{8pt}
\begin{tabular}{l|cc|cc|cc}
\hline
\multirow{2}{*}{} & \multicolumn{2}{l|}{ViT-B-32}      & \multicolumn{2}{l|}{ViT-B-16}      & \multicolumn{2}{l}{Swin-T}        \\ \cline{2-7} 
                  & \multicolumn{1}{c|}{ECE} & KDE-ECE & \multicolumn{1}{c|}{ECE} & KDE-ECE & \multicolumn{1}{c|}{ECE} & KDE-ECE \\ \hline
Uncal. & \multicolumn{1}{c|}{9.53} & 9.56 & \multicolumn{1}{c|}{8.73} & 8.68 & \multicolumn{1}{c|}{9.91} & 9.94 \\
Ours   & \multicolumn{1}{c|}{\textbf{7.55}} & \textbf{7.66} & \multicolumn{1}{c|}{\textbf{8.45}} & \textbf{8.59} & \multicolumn{1}{c|}{\textbf{8.90}} & \textbf{8.89} \\ \hline
\end{tabular}
\end{table}

Transformer-based vision models, including \cite{dosovitskiy2020image} and \cite{liu2021swin}, are known to be relatively well-calibrated, as noted in \cite{minderer2021revisiting}. However, to verify the calirbation efficacy of our method on these transformer-based models, we conducted additional experiments. Table \ref{table:vit} shows the results before and after applying our method to ViT-B-32, ViT-B-16, and Swin-T, respectively. Similarly, these results represent the averaged ECE and KDE-ECE across all levels of corruption severity from no corruption (ID) to maximum corruption severity in ImageNet-C. It is evident from these findings that our method can consistently achieve calibration effects on transformer-based vision models as well. Furthermore, as detailed in Fig.\ref{fig:vit}, our method consistently exhibits calibration effects at every level of corruption severity, demonstrating its robustness.

\section{Evaluation on Additional Covariate OOD Dataset}
\label{sec:cov}
\vspace{-5mm}

\begin{table*}[ht!]
\centering
\caption{\textbf{ECE results on additional covariate shifted datasets. } This table presents ECE using accuracy-preserving methods. It confirms that our proposed method exhibits the best calibration performance across ImageNet-R, ImageNet-A, and ImageNet-Sketch. }
\label{table:more cov}
\centering
\setlength{\tabcolsep}{6pt}
\begin{adjustbox}{width=0.9\textwidth}
\renewcommand{\arraystretch}{1.0}
\begin{tabular}{@{~}c|c|ccccccc@{~}}
 \dtoprule
  \multicolumn{7}{c}{ ImageNet-R }\\ \hline
Network         & Uncalibrated  & TS        & ETS            & IRM             & SPLINE      & \cellcolor{gray!10}Ours \\ \hline
DenseNet201     & 26.84         & 24.35     & 24.35          & 26.28           & 26.37       & \cellcolor{gray!10}\textbf{15.22} \\
ResNet18        & 22.99         & 19.76     & 19.76          & 22.19           & 22.37       & \cellcolor{gray!10}\textbf{11.81} \\
VGG16           & 22.92         & 17.64     & 21.28          & 21.99           & 21.17       & \cellcolor{gray!10}\textbf{9.23} \\
WideResNet40    & 28.87         & 22.91     & 22.45          & 27.13           & 25.18       & \cellcolor{gray!10}\textbf{10.94} \\
 \toprule
  \multicolumn{7}{c}{ ImageNet-A }\\ \hline
DenseNet201     & 36.24         & 33.77     & 33.77          & 35.72           & 35.81      & \cellcolor{gray!10}\textbf{24.81} \\
ResNet18        & 33.42         & 30.31     & 30.31          & 32.69           & 32.80      & \cellcolor{gray!10}\textbf{23.14} \\
VGG16           & 30.76         & 25.67     & 29.20          & 29.77           & 29.01      & \cellcolor{gray!10}\textbf{17.32} \\
WideResNet40    & 36.34         & 30.46     & 30.04          & 34.74           & 32.63      & \cellcolor{gray!10}\textbf{18.67} \\
 \toprule
   \multicolumn{7}{c}{ ImageNet-Sktech }\\ \hline
DenseNet201     & 28.07         & 25.62     & 25.62          & 27.60           & 27.68       & \cellcolor{gray!10}\textbf{17.13} \\
ResNet18        & 21.94         & 18.83     & 18.83          & 21.19           & 21.32       & \cellcolor{gray!10}\textbf{11.53} \\
VGG16           & 22.30         & 17.28     & 20.80          & 21.49           & 20.60       & \cellcolor{gray!10}\textbf{10.15} \\
WideResNet40    & 30.58         & 24.72     & 24.32          & 29.28           & 27.02       & \cellcolor{gray!10}\textbf{13.61} \\ \dbottomrule
\end{tabular}
  \end{adjustbox}
\end{table*}

In the main text, we utilized the ImageNet-C \cite{hendrycks2019robustness} dataset, which features progressively diverging stages of corruption from the in-distribution (ID), to represent covariate-shifted test sets and effectively demonstrate the characteristics of our method. In this section, we introduce additional experimental results using additional covariate OOD datasets as test sets beyond ImageNet-C. We employed different covariate OOD datasets including ImageNet-Renditions (R) \cite{hendrycks2021many}, ImageNet-Adversarial (A) \cite{hendrycks2021natural}, and ImageNet-Sketch \cite{wang2019learning}.

Firstly, ImageNet-R consists of a test set of 30,000 images that encompass a variety of artistic renditions (\eg paintings, graffiti, and embroidery), covering 200 classes from the ImageNet-1k. Secondly, ImageNet-A contains 7,500 natural adversarial examples that are real-world adversarially filtered images designed to challenge existing ImageNet-1k classifiers. Lastly, ImageNet-Sketch, similar to the aforementioned datasets, is used to evaluate generalization and robustness in situations with distribution shifts, consisting of 50,000 sketch-like images from the entire ImageNet-1k classes.

\section{Ablation Study on Semantic OOD Data}
\label{sec:ablation}
This section provides interesting detailed examinations through various ablation studies related to the semantic OOD dataset, which was utilized for parameter tuning in our method. Firstly, Section \ref{sec:ablation_type} discusses the impacts of different types of semantic OOD datasets utilized for tuning.
Lastly, Section \ref{sec:ablation_without} offers a comprehensive analysis of the results when semantic OOD is not used for tuning at all.

\subsection{Semantic OOD type}
\label{sec:ablation_type}
\vspace{-5mm}

\begin{table}[h!]
\centering
\caption{\textbf{Results when using different tuning semantic OOD datasets.} With the use of different semantic OOD datasets for tuning, the calibration performance remains consistently good without significant changes in performance.}
\label{table:ood type}
\setlength{\tabcolsep}{6pt}
\begin{adjustbox}{width=0.9\textwidth}
\begin{tabular}{ll|c|c|c|c|c|c}
\dtoprule
\multicolumn{2}{l|}{Tune ID} & \multicolumn{2}{l|}{CIFAR10} & \multicolumn{2}{l|}{CIFAR100} & \multicolumn{2}{l}{ImageNet-1k} \\ \hline
\multicolumn{2}{l|}{Tune OOD}     & SVHN  & Texture & SVHN  & Texture & SVHN & Texture \\ \hline
\multicolumn{1}{l|}{\multirow{3}{*}{Method}} & Uncal. & 12.90 & 12.90   & 22.79 & 22.79   & 7.32 & 7.32    \\
\multicolumn{1}{l|}{} & SoTA   & 5.78  & 5.78    & 7.52  & 7.52    & 5.57 & 5.57    \\
\multicolumn{1}{l|}{} & Ours   & \textbf{4.47}  & \textbf{4.51}    & \textbf{5.94}  & \textbf{5.88}    & \textbf{5.47} & \textbf{5.31}   \\ \dbottomrule
\end{tabular}
\end{adjustbox}
\end{table}

For the ID dataset, we used CIFAR10, CIFAR100, and ImageNet-1k, while for the semantic out-of-distribution (OOD) datasets, we utilized SVHN and the Texture dataset. When tuning the calibration parameter, if SVHN was used, then another dataset, Texture, was employed as the test set to prevent any data leakage to the test dataset. We highlight that the semantic OOD data used for tuning was not used as the test dataset. In this section, we conducted an ablation study on the types of semantic OOD used for parameter tuning, using DenseNet201. Table \ref{table:ood type} indicates that regardless of the semantic OOD used for tuning the parameters of our method, it consistently maintained high performance without significant changes in the results.

\subsection{Analysis of Results Without Using Semantic OOD for Tuning}
\label{sec:ablation_without}

\begin{table*}[h!]
\centering
\caption{\textbf{Averaged ECE over all severity levels of corruption, including results without using semantic OOD for tuning. } The lowest ECE results were highlighted in bold, and the second-best results were underlined, with the parentheses indicating the semantic OOD dataset used for tuning our method. `Ours w/o' indicates the case where our method does not use semantic OOD for tuning. In many cases, while the performance without using semantic OOD for tuning is somewhat lower than the originally proposed method, it still exhibits better calibration performance compared to other methods. }
\vspace{-2mm}
\label{table:overall without ECE}
\centering
\setlength{\tabcolsep}{4pt}
\begin{adjustbox}{width=\textwidth}
\renewcommand{\arraystretch}{0.95}
\begin{tabular}{@{~}c|c|c|cccccccc@{~}}
 \toprule
 \multicolumn{11}{c}{ECE ($\downarrow$)}\\ \hline
 ID (OOD)&  Network        & Uncalibrated & TS    & ETS   & IRM   & IROvA & IROvATS & SPLINE & \cellcolor{gray!10}Ours w/o & \cellcolor{gray!10}Ours          \\
\hline
\multirow{4}{*}{\begin{tabular}[c]{@{}c@{}}CIFAR10\\ (SVHN)\end{tabular}}        & DenseNet201 & 12.90 & 9.30  & 8.95 & 5.78 & 6.76  & 6.05 & 7.40 & \cellcolor{gray!10}\underline{5.44} & \cellcolor{gray!10}\textbf{4.47} \\
 & VGG19        & 16.17 & 10.59 & 10.34 & 7.18  & 7.92  & 7.33    & 7.76  & \cellcolor{gray!10}\underline{6.53}    & \cellcolor{gray!10}\textbf{5.78} \\
 & WideResNet40 & 12.49 & 8.57  & 8.33  & 5.72  & 7.80  & 7.35    & 7.30  & \cellcolor{gray!10}\underline{5.40}    & \cellcolor{gray!10}\textbf{4.71} \\
 & SE-ResNet50  & 14.80 & 10.17 & 9.94  & 8.13  & 8.51  & 8.34    & 9.09  & \cellcolor{gray!10}\underline{7.53}    & \cellcolor{gray!10}\textbf{6.28} \\
 \hline
\multirow{4}{*}{\begin{tabular}[c]{@{}c@{}}CIFAR100\\ (SVHN)\end{tabular}}       & DenseNet201 & 22.79 & 11.21 & 9.73 & 8.14 & 10.88 & 8.02 & 7.52 & \cellcolor{gray!10}\underline{6.91}   & \cellcolor{gray!10}\textbf{5.94} \\
 & VGG19        & 32.49 & 12.54 & 11.99 & 8.65  & 11.42 & 8.55    & \underline{8.23}  & \cellcolor{gray!10}10.45   & \cellcolor{gray!10}\textbf{7.35} \\
 & WideResNet40 & 23.83 & 13.32 & 11.17 & 10.50 & 13.88 & \underline{9.69}    & 9.90  & \cellcolor{gray!10}10.12   & \cellcolor{gray!10}\textbf{8.74} \\
 & SE-ResNet101 & 30.27 & 15.24 & 12.89 & \underline{10.17} & 13.59 & 10.67   & 10.46 & \cellcolor{gray!10}10.45   & \cellcolor{gray!10}\textbf{8.95} \\
 \hline
\multirow{2}{*}{\begin{tabular}[c]{@{}c@{}}ImageNet-1k\\ (Texture)\end{tabular}} & DenseNet201   & 6.99  & \underline{5.49}  & \underline{5.49} & 6.37 & 10.67 & 8.95 & 6.44 & \cellcolor{gray!10}\underline{5.49} & \cellcolor{gray!10}\textbf{4.92} \\
 & ResNet18  & 7.32  & \underline{5.57}  & \underline{5.57}  & 6.77  & 9.58 & 8.06    & 6.83  & \cellcolor{gray!10}\underline{5.57}    & \cellcolor{gray!10}\textbf{5.31}\\ \hline
 
\end{tabular}
  \end{adjustbox}
\end{table*}

Our intuition suggests that using energy to distinguish between correct and incorrect predictions should be effective with our algorithm alone. However, exposure to semantic OOD, which is drastically different from the train distribution, likely enhances robust calibration against OOD scenarios. To investigate this further, we analyzed the results when not using semantic OOD data for tuning at all. Under the same experimental settings as in the main text, we configured and tested our method without using semantic OOD data, adding these results for comparative analysis.

\begin{figure}[h!]
\centering
    \includegraphics[width=1.0\columnwidth]{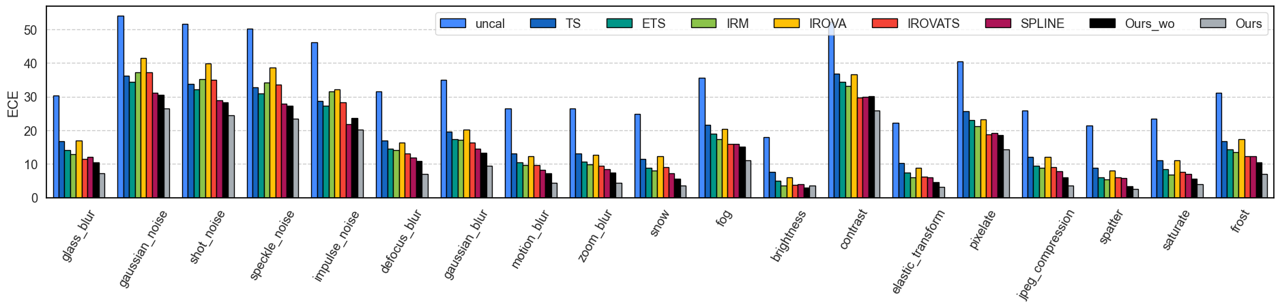}
\caption{\textbf{ECE for various corruption types at severity level 5, including results without using semantic OOD for tuning.  } For most types of corruption, the performance is somewhat lower than that of the original proposed method but still better compared to other methods. }
\label{fig:without ctype bar}
\end{figure}

\begin{figure}[h!]
\centering
    \includegraphics[width=1.0\columnwidth]{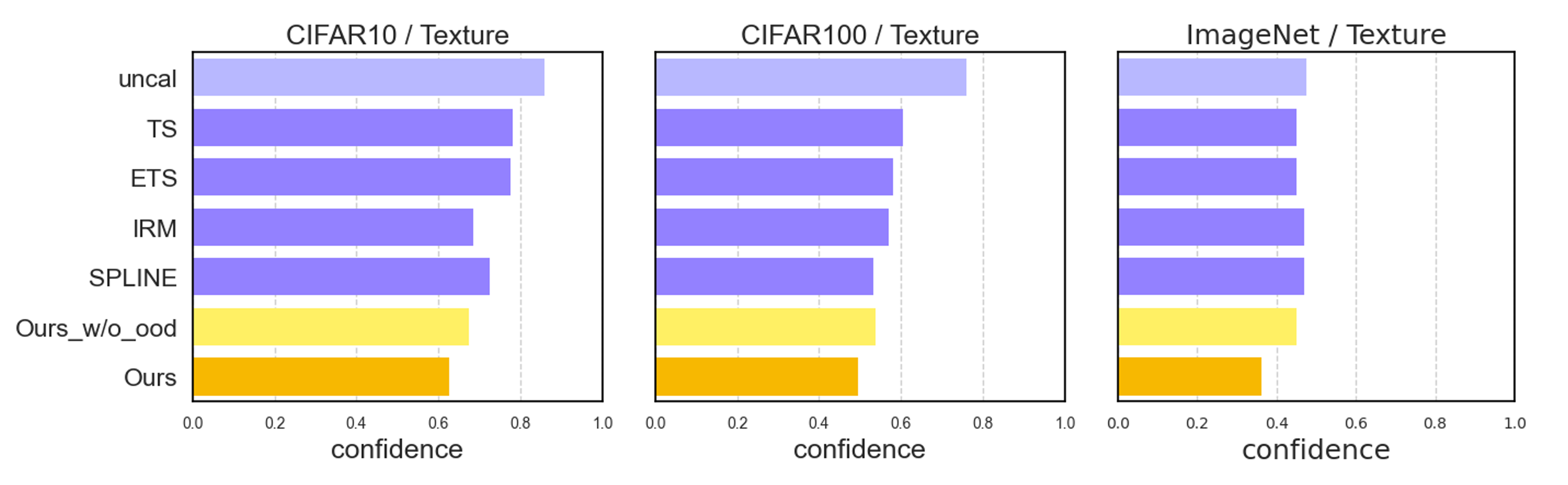}
\caption{\textbf{Calibrated confidence scores for semantic OOD samples, including results without using semantic OOD for tuning.  } Each bar denotes the confidence score on semantic OOD samples. A shorter bar indicates better-calibrated confidence. For this experiment, we used DenseNet201. }
\label{fig:without semantic conf}
\vspace{-3mm}
\end{figure}

Firstly, when comparing the `Ours w/o' results in Table \ref{table:overall without ECE}, where semantic OOD was not used, it interestingly shows better performance in many cases compared to other state-of-the-art methods, albeit less effective than our full proposed method utilizing semantic OOD. In more details, as shown in Fig.\ref{fig:without ctype bar}, the DenseNet201(CIFAR100) results according to the corruption type at the maximum corruption level, while not as good as our original method, still outperform other state-of-the-art methods for most corruption types. Therefore, this demonstrates that the results consistent with our intuition and the intended design of our method. Lastly, consistent results were also observed for the semantic OOD test set. As shown in Fig.\ref{fig:without semantic conf}, although the results without using the semantic OOD dataset for tuning are less effective compared to our original method, they still show the best calibration effect compared to other existing methods.